%% file: aaai22.tex
\def\year{2022}\relax
\documentclass[letterpaper]{article} 
\usepackage{aaai22}  
\usepackage{times}  
\usepackage{helvet}  
\usepackage{courier}  
\usepackage[hyphens]{url}  
\usepackage{graphicx} 
\usepackage{natbib}  
\usepackage{caption} 
\DeclareCaptionStyle{ruled}{labelfont=normalfont,labelsep=colon,strut=off} 
\usepackage[linesnumbered,algoruled,boxed,lined,noend]{algorithm2e}

\newcommand{\modelName}{\textsc{CoRGi}\xspace}

\usepackage{newfloat}
\usepackage{listings}
\nocopyright

\usepackage{subcaption}
\usepackage{wrapfig}
\usepackage{booktabs}
\usepackage{multirow}
\usepackage{amssymb}
\usepackage{amsmath}
\usepackage{bm}
\usepackage{pifont}
\usepackage{paralist}
\newcommand{\cmark}{\ding{51}}%
\newcommand{\xmark}{\ding{55}}%

\title{\modelName: Content-Rich Graph Neural Networks with Attention}

\author {
    Jooyeon Kim,\textsuperscript{\rm 1}\footnote{Work done while at Microsoft Research.}
    Angus Lamb, \textsuperscript{\rm 2}
    Simon Woodhead, \textsuperscript{\rm 3}
    Simon Peyton Jones, \textsuperscript{\rm 2}
    Cheng Zheng, \textsuperscript{\rm 2}
    Miltiadis Allamanis \textsuperscript{\rm 2}
}
\affiliations {
    \textsuperscript{\rm 1} RIKEN, Japan\\
    \textsuperscript{\rm 2} Microsoft Research, Cambridge, UK\\
    \textsuperscript{\rm 3} Eedi, UK\\
}

\begin{document}

\maketitle
\begin{abstract}
Graph representations of a target domain often project it to a set of entities (nodes) and their relations (edges). However, such projections often miss important and rich information. For example, in graph representations used in missing value imputation, items --- represented as nodes --- may contain rich textual information. However, when processing graphs with graph neural networks (GNN), such information is either ignored or summarized into a single vector representation used to initialize the GNN. Towards addressing this, we present \modelName, a GNN that considers the rich data within nodes in the context of their neighbors. This is achieved by endowing \modelName's message passing with a personalized attention mechanism over the content of each node. This way, \modelName assigns \emph{user-item-specific} attention scores with respect to the words that appear in an item's content. We evaluate \modelName on two edge-value prediction tasks and show that \modelName is better at making edge-value predictions over existing methods, especially on sparse regions of the graph.
\end{abstract}
  
  \input{1_introduction}
  \input{3_methodology}
  \input{2_related_work}

  \input{4_experiments}
  \input{5_conclusion}

\clearpage

\bibliography{ref}

\clearpage

\appendix
\input{appendix}
\end{document}

%% file: 1_introduction.tex
\section{Introduction}
Graph neural networks (GNN) have enjoyed great success in
deep learning. GNNs allow us to model complex graph-structured data. However, the construction of the input graphs is often a lossy projection of the data of the modeled domain.
For example, a graph representation of a book recommendation problem
may represent books and users as nodes with valued edges as recommendations.
However, each book node contains rich semi-structured content,
such as text structured into sections, tables, etc., which can be used to improve the performance of recommendations.

A common approach to incorporate node content in GNNs
is to ``summarize'' it into a single vector representation (embedding) and use the vector as an initial node embedding. This often includes computing a single vector representation from a whole sentence or document using an encoder model, such as a bag-of-words model or a transformer. However, such
representations are suboptimal, given the relatively small size of these vectors compared to the original content.
This is widely accepted in
natural language processing (NLP), and instead
of representing inputs as a single vector/embedding, the full input is used. For example, encoder-decoder models employ some form of attention mechanism over the whole input given the model context instead of representing it as a single vector. For example, in NLP text summarization \citep{you2019improving}, a decoder attends
to the encoded representations of all the words in the input text.

In the same fashion, we need a better way for a GNN to capture the content within
nodes of a graph. Towards this goal, we present \modelName (\underline{Co}ntent-\underline{R}ich \underline{G}raph neural network with attent\underline{i}on), a message-passing GNN~\citep{gilmer2017neural} that incorporates an attention mechanism
over the rich \emph{node} content during each message passing step. This allows \modelName
to effectively learn \emph{both} about the structure
of the graph and the content within each node.

\begin{figure*}
    \centering
    \includegraphics[width=0.6\textwidth]{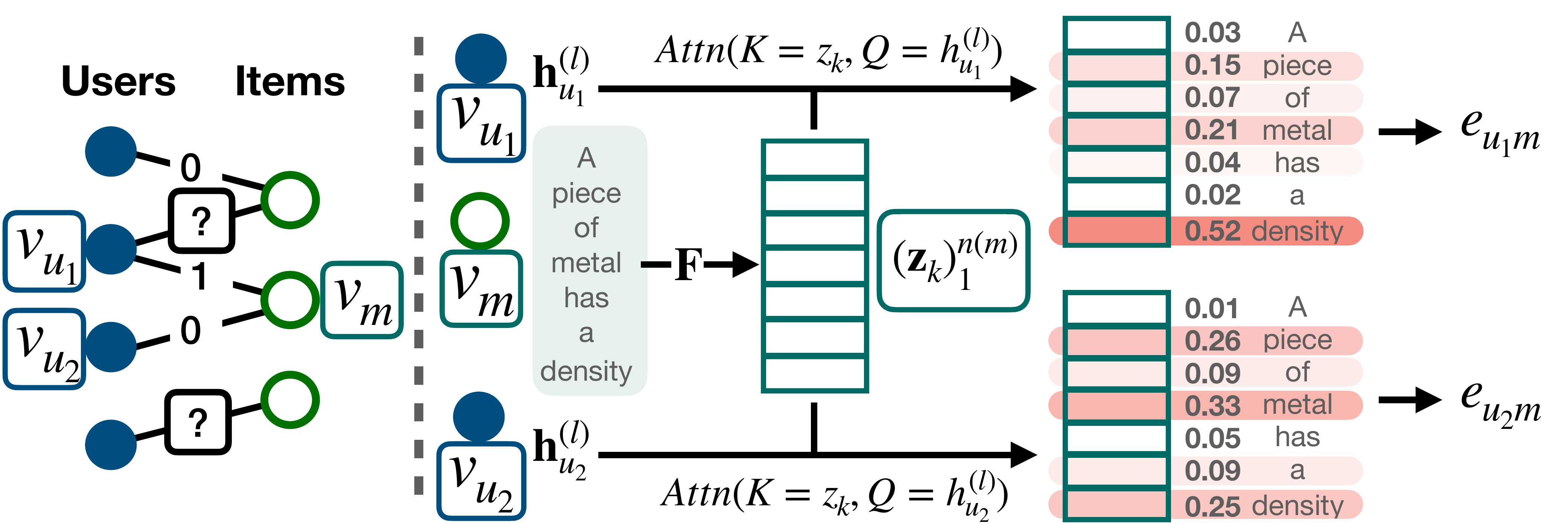}
    \caption{
        \textbf{Left}: In an educational setting, students (users) and questions (items) form a bipartite graph. Predicting student responses can be posed as missing edge value imputation. Edge values are known for some user-item pairs but not for all. \textbf{Right}: The edge representation computation in \modelName's $l$th message-passing layer. Item nodes are associated with content (e.g., a question node $\mathit{v}_m$ contains text). The node content is encoded by a model $\mathbf{F}$ (e.g., a transformer), obtaining a set of content vectors $\mathbf{Z_m}=\{\mathbf{z}_k\}_1^{n(m)}$. During message passing, \modelName computes the representation of an edge (e.g., $\mathbf{e}_{u_1m}$) using the last layer's node embedding of a user (student) node (e.g., $\mathbf{h}_{u_1}^{(l)}$) and an attention mechanism over $\mathbf{Z_m}$. Thus, message passing takes into consideration the full content within the item (question) nodes in a personalized user-dependent way.
    }
    \label{fig:intro}
\end{figure*}

One interesting application of \modelName
is edge-value imputation, e.g., missing value imputation with GNNs in collaborative filtering~\citep{you2020handling} (Fig.~\ref{fig:intro}-left).
For example, in a dataset of
student-question answers, each question is associated with a rich
textual description. In such settings, graph-based
representations capture the rich interactions among students and questions (user responses) but ignore important content within items (textual descriptions of questions; text in Fig.~\ref{fig:intro}).
\modelName combines both sources of information through a personalized attention-based message passing method that computes user-item (student-question) pair-specific  representations of the item's content.
Additionally, \modelName achieves better performance compared to baselines, particularly
in sparse regions of the user-item graph, such
as rarely answered questions. 

\paragraph{Contributions} In summary, our contributions are
\begin{inparaitem}
    \item \modelName: a message-passing GNN that incorporates
    an attention mechanism over node content during messages computation.
    \item A specialization of \modelName for the user-item recommendation.
    \item An extensive evaluation over two real-world datasets showing that \modelName improves user-response prediction performance over existing methods, especially for items with few user
    ratings or in sparse graphs, where content plays a critical role.
\end{inparaitem}

%% file: 3_methodology.tex
\section{\modelName Architecture}\label{sec:method}
In this section, we first describe  \modelName's problem setting
and its implementation. We then focus on the recommender systems use-case with textual content in item nodes.
Finally, we discuss \modelName's computational complexity and how to reduce it.
The appendix summarizes our notation.
\begin{algorithm}[tb]
\SetAlgoLined
\For{$\textit{l} \in \{1, \dots , L \}$  } {
    \For{$\textit{i} \in \mathcal{V}$}{
        $\mathbf{m}_{ij}^{(l)} \leftarrow$ ~\text{Compute messages from} Eq.~\ref{eqn:message} \;
        $\mathbf{h}_{\mathit{i}}^{(l)} \leftarrow$ ~Update node states with Eq.~\ref{eqn:node_embedding} \;
        %
    
    %
    %
    \For{$v_j \in \mathcal{N}(i)$}{
        $\mathit{c}_{ik}^{(l)} \leftarrow$ Compute CA scores from Eq.~\ref{eqn:attn_coef1} or Eq.~\ref{eqn:attn_coef2} \;
        $\alpha_{ik}^{(l)} \leftarrow$ Compute content attention probability from Eq.~\ref{eqn:attn_prob} \; 
        $\mathbf{e}_{ij}^{(l)\prime} \leftarrow$ Update content-independent edge embedding with Eq.~\ref{eqn:edge_emb_grape} \;
        $\mathbf{e}_{ij, \text{CA}}^{(l)} \leftarrow$ Update content-attention edge embedding with Eq.~\ref{eqn:edge_emb} or use cache\;        
        $\mathbf{e}_{ij}^{(l)} \leftarrow \mathbf{e}_{ij}^{(l)\prime} + \mathbf{e}_{ij, \text{CA}}^{(l)}$
    }
}
}

\vspace{2mm}
\KwOut{Node embeddings $\mathbf{h}_{ \mathit{v} }, \forall \mathit{v} \in \mathcal{V} $}
\caption{\modelName message-passing computation}
\label{alg:main}
\end{algorithm}
\paragraph{Problem setting}
Consider a graph $\mathcal{G}= (\mathcal{V}, \mathcal{E})$. Each node $\mathit{v}\in \mathcal{V}$ is associated with node features $\mathbf{h}_\mathit{v}^{(0)}$ and each edge with the features $\mathbf{e}_{ij}^{(0)}, \forall (i, j) \in \mathcal{E}$.
If a node or edge is \emph{not} related with any features, a constant value may be assigned.
A subset of nodes $\mathcal{V}_{C} \subset \mathcal{V}$ is associated with a set of
$n(\mathit{i})$ \emph{content} vector representations $\mathbf{Z}_{\mathit{i}} = \{ \mathbf{z}_1^{(\mathit{i})}, \dots , \mathbf{z}^{(\mathit{i})}_{n(\mathit{i})}\}, \forall v_i \in \mathcal{V}_{C}$, and $\mathbf{z}_k^{(\mathit{i})} \in \mathbb{R}^{\mathit{D}}$. 
Notice that $n (\mathit{i})$, the number of content vectors of each node $v_i$, may differ.
$\textbf{Z}_i$ may be given or computed with
a  node content encoder (e.g., a transformer). 
\paragraph{\modelName} The goal
is to learn representations over the nodes while considering both the graph structure and the set of content vectors within each node (Fig.~\ref{fig:intro}). \modelName follows
the message-passing GNN paradigm~\citep{gilmer2017neural} and is closely related to GRAPE~\citep{you2020handling}.
In contrast to existing models, \modelName uses the content vector representations associated with each node during message-passing with personalized attention.
Specifically, \modelName computes messages by learning to focus on potentially different parts of the content in the context of the neighboring nodes using an attention mechanism (Fig.~\ref{fig:intro}).
Alg.~\ref{alg:main} presents a high-level overview of \modelName, discussed next.

\modelName's architecture assumes $L$ message passing layers.
Following \citet{you2020handling},
at each layer $l$, \modelName computes a \emph{message} $\mathbf{m}_{\mathit{ij}}^{(l)}$ from node $\mathit{v}_j$ to $\mathit{v}_i$ using the previous-level node embedding $\mathbf{h}_{i}^{(\mathit{l}-1)}$ and edge embedding $\mathbf{e}_{ij}^{(l-1)}$ as
\begin{equation}
    \mathbf{m}_{\mathit{ij}}^{(l)} =   \sigma \left( \mathbf{P}^{(l)} \cdot \left[ \mathbf{h}_{\mathit{j}}^{(\mathit{l} - 1)}, \mathbf{e}_{ij}^{(l-1)}\  \right]  \right) ,
    \label{eqn:message}
\end{equation}
where $\sigma$ is a non-linearity, $[\cdot]$ is vector concatentation, and $\mathbf{P}^{(l)}$ is a trainable weight.
We set $\mathbf{h}_i^{(0)}$ and $\mathbf{e}_{ij}^{(0)}$ to the input node features and input edge attributes (if any),
and $\mathbf{h}^{(l)}\in\mathbb{R}^{\mathit{C}^{(l, h)}}$, $\mathbf{e}^{(l)}\in\mathbb{R}^{\mathit{C}^{(l, e)}}$.
Messages are \emph{aggregated} from all neighbors $\mathcal{N}(i)$ of $v_i$ and node embedding are \emph{computed} as
\begin{equation}
    \mathbf{h}_{i}^{(l)} = \sigma \left(\mathbf{Q}^{(l)} \cdot \left[ \textbf{h}_i^{(l-1)}, \hspace{0.2em} \textsc{Agg}^{(l)} \left(\mathbf{m}_{ij}^{(l)} \hspace{0.2em} \vert  \hspace{0.2em} \forall j \in \mathcal{N}(i) \right)\right]\right),
    \label{eqn:node_embedding}
\end{equation}
where $\textsc{Agg}^{(l)}$ is a permutation-invariant aggregation function, and $\mathbf{Q}^{(l)}$ is a learnable parameter. %

We are interested in incorporating information from the content of each node $\mathit{v}_j\in\mathcal{V}_{C}$. To achieve this,  we use an attention mechanism within the GNN's message-passing. This allows a message between a $\mathit{v}_i$ and $\mathit{v}_j$ to \emph{focus} on a specific part of the content. Such an ability can be helpful in many scenarios.
For example, in an educational recommender system, the attended (textual) content of a question is an essential factor 
in predicting the student's ability to answer it correctly given, e.g., a diagnostic question \cite{wang2020diagnostic}.
Intuitively, a student --- given her skills --- will focus on different aspects of a question when answering it. 

\modelName's attention mechanism aims to emulate this. To model this, we combine any edge features $\mathbf{e}_{ij}^{(0)}$ with a
content-attention vector $\mathbf{e}_{ij, \text{CA}}^{(l)}$ computed from an attention mechanism over the content $\mathbf{Z}_{\mathit{v}_j}$.
We consider two options for this: (a) element-wise addition
$\mathbf{e}_{ij}^{(l)} = \mathbf{e}_{ij}^{(l)\prime} + \mathbf{e}_{ij, \text{CA}}^{(l)}$,
and (b) concatenation
$\mathbf{e}_{ij}^{(l)} = \left[ \mathbf{e}_{ij}^{(l)\prime}, \mathbf{e}_{ij, \text{CA}}^{(l)}\right]$,
where 
\begin{equation}
    \mathbf{e}_{ij}^{(l)\prime} = \sigma \left( \mathbf{W}^{(l)} \cdot \left[\textbf{h}_j^{(l)}, \mathbf{e}_{ij}^{(0)}\right]\right),
    \label{eqn:edge_emb_grape}
\end{equation}
with a trainable weight $\mathbf{W}^{(l)}$, and  $\mathbf{e}_{ij, \text{CA}}^{(l)}$ is computed
by the attention mechanism, discussed next. 
Note that for elementwise addition $\mathbf{e}_{ij}^{(l)\prime}$ must have the same cardinality as $\mathbf{e}_{ij, \text{CA}}^{(l)}$.

Finally, we describe the attention mechanism computing $\mathbf{e}_{ij, \text{CA}}^{(l)}$.
For an edge between $\mathit{v}_i$ and $\mathit{v}_j$ at $l^{th}$ layer, the content-attention (CA) is computed
using the set of content vector representations of $\mathit{v}_j$, $\mathbf{Z}_j$, and the previous-level node embedding $\mathbf{h}^{(l-1)}_{i}$, i.e.,
\begin{align}
\begin{split}
    \mathbf{e}_{ij, \text{CA}}^{(l)} &= \textsc{Attention}\left(\textsc{keys}=\mathbf{Z}_j, \hspace{0.3em} \textsc{query}=\textbf{h}_i^{(l-1)}\right) \\
              &= \sum_k \alpha_{ik} \mathbf{W}_M^{(l)}\mathbf{z}_k^{(\mathit{v}_j)}
    \label{eqn:edge_emb}
\end{split}
\end{align}
where  $\mathbf{W}_M^{(l)}$ is a trainable weight, and
$\alpha_{ik}$ is computed as
\begin{equation}
    \alpha_{ik}^{(l)} = \textsc{SoftMax}_k \left( \mathit{c}_{ik}^{(l)} \enskip \vert \enskip \forall k \in \{ 1, \dots, n(i) \} \right).
    \label{eqn:attn_prob}
\end{equation}
We test two common mechanisms for computing $\mathit{c}_{ik}^{(l)}$: concatenation (CO) and dot-product (DP), computed as 
\begin{align}
    \mathit{c}_{ik, CO}^{(l)} = \sigma \Big ( {\mathbf{p}^{(l)}}^\top  \left[\mathbf{W}_U^{(l)} \mathbf{h}^{(l-1)}_i, \mathbf{W}_M^{(l)} \mathbf{z}_k  \right] \Big )
    \label{eqn:attn_coef1}
\end{align} \begin{align}
    \mathit{c}_{ik, DP}^{(l)} = \sigma \Big ( \left[ \mathbf{W}_U^{(l)}, \mathbf{h}^{(l-1)}_i \right]^\top \mathbf{W}_M^{(l)} \mathbf{z}_k \Big ),
    \label{eqn:attn_coef2}
\end{align}
where $\mathbf{W}_U^{(l)}$, and $\mathbf{p}^{(l)}$ are learnable weights.
Note that the attention is over the node content and should not be confused with
the attention used in GATs~\citep{velivckovic2017graph}. We provide a detailed
explanation in the related work section.
\paragraph{Content representations}
So far, we assumed that the content representation vectors $\mathbf{z}_j^{(\mathit{v}_i)} \in \mathbb{R}^\mathit{D}$
are given. In practice, these representations can be computed from some deep learning component $\mathbf{F}$. \modelName does \emph{not} impose a structure on $\mathbf{F}$. For example if the node content is images, then CNN-based architectures for $\mathbf{F}$ would be reasonable. Similarly,
if the content is text, i.e., a sequence of words (or any other sequence), then any NLP model can be used. This includes text representation models~\citep{bojanowski2017enriching,mikolov2013distributed,pennington2014glove} and sequence encoders~\citep{cho2014learning,peters2018deep,sutskever2014sequence}, including transformers \citep{devlin2019bert,vaswani2017attention}. Such models ``contextualize'' each individual word in the sequence and convert it to a \emph{set} of vector representations.

\subsection{\modelName for user-response prediction} \label{subsec:user response prediction}
User response prediction can be formulated as an edge-value prediction~\citep{berg2017graph,wang2019neural,you2020handling,zhang2019inductive}.
We operationalize \modelName for recommender systems by considering a \emph{bipartite} graph $\mathcal{G} = ( \mathcal{V}, \mathcal{E})$ with two disjoint node sets $\mathcal{V} = \mathcal{V}_U \cup \mathcal{V}_M$ of users and items.
Each item $\mathit{v}_m\in\mathcal{V}_M$ contains text $\mathbf{D}_m = \left[w_1^{(m)}, \cdots, w_{n(m)}^{(m)}\right]$, i.e., a sequence of words.
The sequence is converted to a set of content vectors $\mathbf{Z}_m$ using a sequence encoder and is input to \modelName.
An edge value prediction, i.e., a recommendation  $r(\mathit{v}_m, \mathit{v}_u)$ for a user $\mathit{v}_u$ about an item $\mathit{v}_m$ is made with a read-out layer, defined as
\begin{align}\label{eq:recommendation prediction}
    r(\mathit{v}_m, \mathit{v}_u)=\sigma\left(\mathbf{w}_{\text{out}}^T  \left[\mathbf{h}_u^{(L)}, \mathbf{h}_m^{(L)}\right] + b\right),
\end{align}
where $\mathbf{w}_{\text{out}}$ and $b$ are learnable weights.
\subsection{Complexity analysis}
\label{subsection:complexity}
\modelName's computational and memory complexity is similar to most 
message-passing GNNs, with the additional cost of the attention mechanism. Compared to the node-to-node attention of GATs, \modelName's attention mechanism involves maximum of $T = \max_{v\in\mathcal{V}_M} ( \vert \mathbf{Z}_v \vert )$, the maximum content size with respect to $v \in \mathcal{V}_C$, for each content node.
For the $l^{th}$ message-passing \modelName layer the computational complexity for computing the content attention is expressed as:
\begin{align}
\begin{split}
\mathcal{O} \Big (
    \vert \mathcal{V}_U \vert \cdot \mathit{C}^{(l-1, h)} \cdot \mathit{C}^{(l, e)}
    &+
    \vert \mathcal{V}_M \vert \cdot \mathit{T} \cdot \mathit{D} \cdot \mathit{C}^{(l, e)}\\
    &+
    \vert \mathcal{E}\vert \cdot \mathit{T} \cdot \mathit{C}^{(l, e)}
\Big ),
\end{split}
\end{align}
The first and the second terms arise from the multiplication between the trainable weights and the node embeddings or content vector representations, in Eq.~\ref{eqn:attn_coef1} or ~\ref{eqn:attn_coef2}.
The last term is due to the pairwise linear operation in the attention coefficient calculation between the queries and keyes, in Eq.~\ref{eqn:edge_emb}.

We can drastically reduce the complexity by using the neighbor sampling method proposed by~\citet{hamilton2017inductive}
and caching all $\mathbf{e}_{ij, \text{CA}}^{(l)}$.
For network sampling, we sample a subset of nodes $\mathcal{V}^\prime = \big\{ \mathcal{V}_{U}^\prime \cup \mathcal{V}_{M}^\prime \big\}$ for the neighbor sampling and only update $\mathbf{e}_{ij, \text{CA}}$ whose target node $v_j$ is in $\mathcal{V}_{M}^\prime$ and source node $v_i$ is in $\mathcal{N}(\mathcal{V}_{M}^\prime)$.
The sampled subgraph is $\mathcal{G}^\prime = \big\{ \mathcal{V}^{\prime\prime}, \mathcal{E}^\prime \big\}$, with $\mathcal{V}^{\prime\prime} = \big\{ \mathcal{V}_M^\prime \cup \mathcal{N}(\mathcal{V}_M^\prime) \big\}$.
This way, the computational cost is reduced.

Computing the attention for each layer is costly both in
terms of memory and computation. To drastically reduce the memory and compute requirements, we use a caching trick for all $\mathbf{e}_{ij, \text{CA}}^{(l)}$. Since these representations
can be thought as edge features, we want to compute them infrequently
and re-use them. To do this, we create a cache for all $\mathbf{e}_{ij, \text{CA}}^{(l)}$ and initialize them with zeros. Then,
at the final layer $L$, we compute $\mathbf{e}_{ij, \text{CA}}^{(L)}$  using Eq.~\ref{eqn:edge_emb} and
update the cache for all $\mathbf{e}_{ij, \text{CA}}^{(l)}$ to
the computed $\mathbf{e}_{ij, \text{CA}}^{(L)}$. The newly cached values will be used in subsequent message-passing iterations.
In this way, we avoid $L-1$ computations of Eq.~\ref{eqn:edge_emb}.

%% file: 2_related_work.tex
\section{Related work}
\label{section:related_work}

\modelName is at the intersection of GNNs and machine learning models for missing value imputation. 
Related work that is used as baselines in the evaluation is \emph{emphasized}.

Missing value imputation is the task of filling in previously unknown entries with predicted values.
For two heterogeneous groups, namely, users and items, the task is commonly reduced to matrix completion~\citep{bennett2007netflix} with numerous collaborative filtering and matrix factorization approaches~\citep{billsus1998learning,koren2009matrix,linden2003amazon,mcauley2013hidden,mnih2007probabilistic,sarwar2001item} and deep learning-based approaches~\citep{spinelli2020missing,vincent2008extracting,yoon2018gain}.
\emph{Deep matrix factorization (DMF)}~\citep{xue2017deep} directly uses the input matrix by feeding this information through multilayer perceptrons (MLPs).
An extension to variational autoencoders (VAE)~\citep{kingma2013auto}, the \emph{partial-VAE model (PVAE)}~\citep{ma2019eddi} is an encoder-decoder-based approach for imputing missing values.
In contrast, \modelName leverages additional content information in a user-item-specific manner through attention in message-passing.

Over the past years, there have been attempts to model graphs
\citep{grover2016node2vec,perozzi2014deepwalk,tang2015line}.
\citet{kipf2016semi} proposed \emph{graph convolutional networks (GCNs)}, a neural network that learns latent representations of nodes, amongst other deep neural network-based approaches~\citep{bruna2013spectral,defferrard2016convolutional,duvenaud2015convolutional,li2015gated,niepert2016learning,scarselli2008graph}.
\emph{GraphSAGE}~\citep{hamilton2017inductive} extends GCNs allowing the model to be trained on some part of the graph, enabling inductive learning settings.
\emph{Jumping knowledge (JK)} network~\citep{xu2018representation} and \emph{Graph Isomorphism Network (GIN)}~\citep{xu2018powerful} are proposed to improve the representation power of GNNs by adopting new aggregation schemes with respect to the representations of different layers and their previous representations.
A GNN model that is designed for recommender systems, \emph{graph convolutional matrix completion (GC-MC)}~\citep{berg2017graph} is a variant of GCNs that explicitly uses edge labels as inputs to model messages.
Compared to other approaches, GC-MC employs a single-layer message-passing scheme, and each label is endowed with a separate message passing channel.
\emph{GRAPE}~\citep{you2020handling} employs edge embeddings on GCNs
and adopts edge dropout applied throughout all message-passing layers.
LightGCN~\citep{he2020lightgcn} designs a GCN framework that simplifies or omits constructions that are not beneficial for recommendations, such as feature transformation and nonlinear activation, and puts more emphasis on neighborhood aggregation.
Compared to the previously proposed GNN models in recommender systems, \modelName leverages the rich content information of nodes to model a target domain projected to graphs.
Compared to existing GNN models that exploit the content information of nodes,
\modelName employs an attention mechanism over node content within the GNN's message-passing and
computes user-item-specific attention which is used to update edge embeddings. 
\citet{wu2020graph} surveys the recent literature on GNNs for recommender systems.

In natural language processing, the (self-)attention is used to relate word or word tokens of a given sequence~\citep{lin2017structured,parikh2016decomposable,paulus2018deep,vaswani2017attention}.
Many GNN models~\citep{gao2019graph,hou2019measuring,kim2021how,zhang2018gaan}, with \emph{graph attention networks (GATs)}~\citep{velivckovic2017graph} being a popular example, use an attention mechanism to allow the target nodes to distinguish the weights of multiple messages from the source nodes for aggregation.
We note that \modelName is orthogonal to the GAT-like models; although \modelName uses attention, it is over the content within each node instead of the neighbors of each node (as in GATs). In future work \modelName-like mechanisms can be embedded to GAT-like GNNs.

%% file: 4_experiments.tex
\section{Evaluation}
\label{sec:experiments}
\paragraph{Model configuration} 
We employ the $L=3$ \modelName with node embedding and edge embedding cardinalities $\mathit{C}^{(l, h)}$ and $\mathit{C}^{(l, e)}, \forall l \in \{ 1, \dots, L\}$  set to 64,
and size of the prediction read-out layer in Eq.~\ref{eq:recommendation prediction} is set to 256.
We initialize node embeddings to random values, and assign the train label values to initialize the edge embeddings.
We use mean pooling for aggregation.
For the non-linear activation, we use LeakyReLU for attention coefficient computation (Eq.~\ref{eqn:attn_coef1},~\ref{eqn:attn_coef2}) with negative slope set to 0.2, as suggested by~\citet{velivckovic2017graph} and ReLU~\citep{nair2010rectified} for the rest.

\paragraph{Training configuration}
For all experiments, we train \modelName 
with Adam~\citep{kingma2014adam} and a learning rate of 0.001.
We employ early stopping on validation loss, with train, test, and validation sets split in 8:1:1 ratio.
We use binary cross entropy loss (BCE) for binary values and mean squared error (MSE) for ordinal values.
We apply dropout~\citep{srivastava2014dropout} on the message passing layers, the prediction MLPs, as well as on edges, with rates chosen from $\{$0.1, 0.3, 0.5, 0.7$\}$ with respect to the validation set performance.
For the baselines, the parameter settings are done in the following manner: 
1) When the settings of the comparison models overlap with \modelName's, e.g., the number of message passing layers or the learning rate, we used the same configurations as \modelName.
2) For the parameter settings that are unique to the comparison model, we followed the setting that is disclosed in the original paper.
3) When the setting disclosed in the original paper is not applicable to the datasets used or our training environment, we select those that yield the best validation performance.
\renewcommand{\arraystretch}{1}
\begin{figure*}[t]
  \centering
  \begin{subfigure}[b]{0.19\textwidth}
    \centering
    \includegraphics[width=\textwidth]{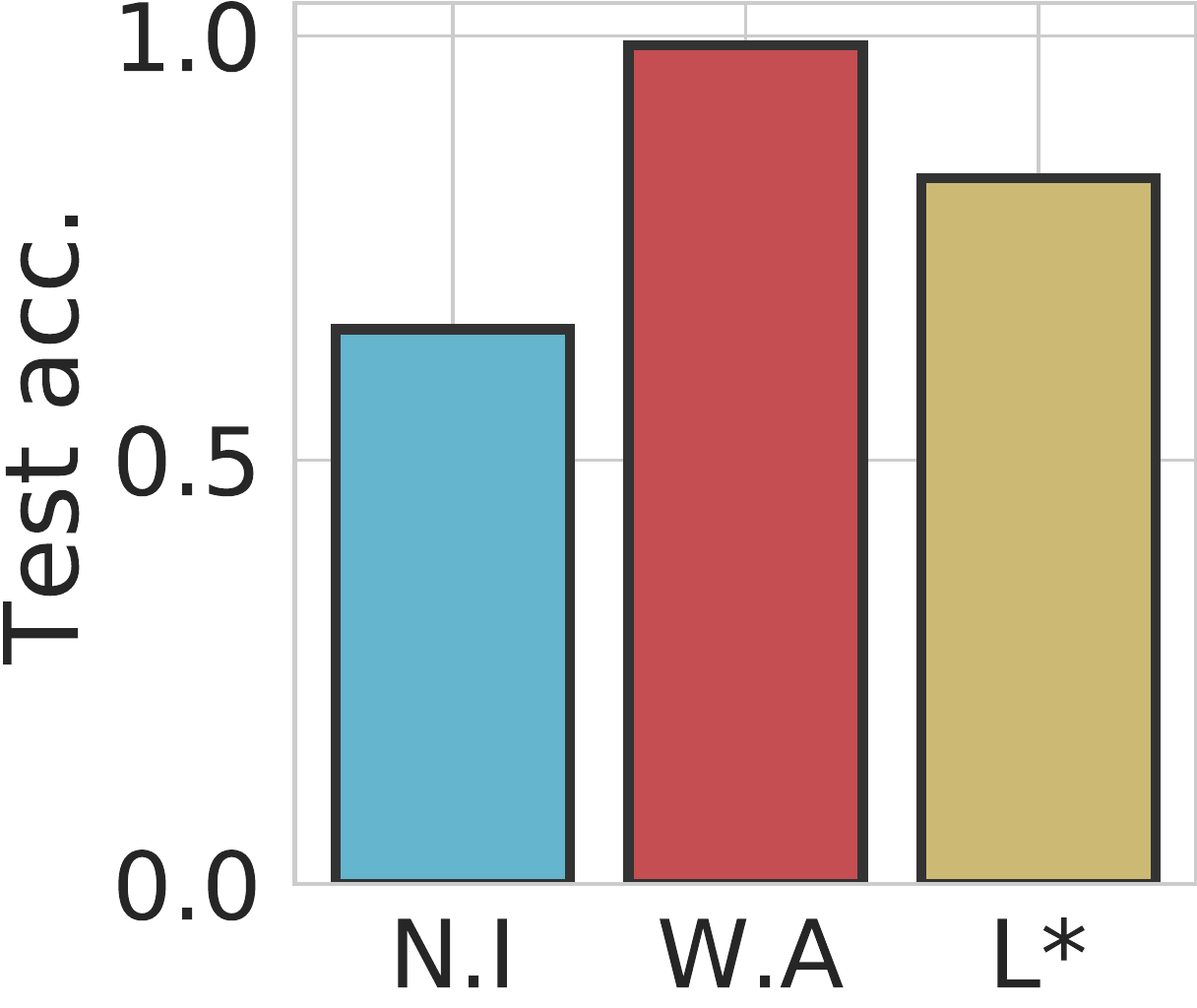}
    \caption{}
    \label{subfig:synthetic_test_acc}
  \end{subfigure}
  %
  \begin{subfigure}[b]{0.195\textwidth}
    \centering
    \includegraphics[width=\textwidth]{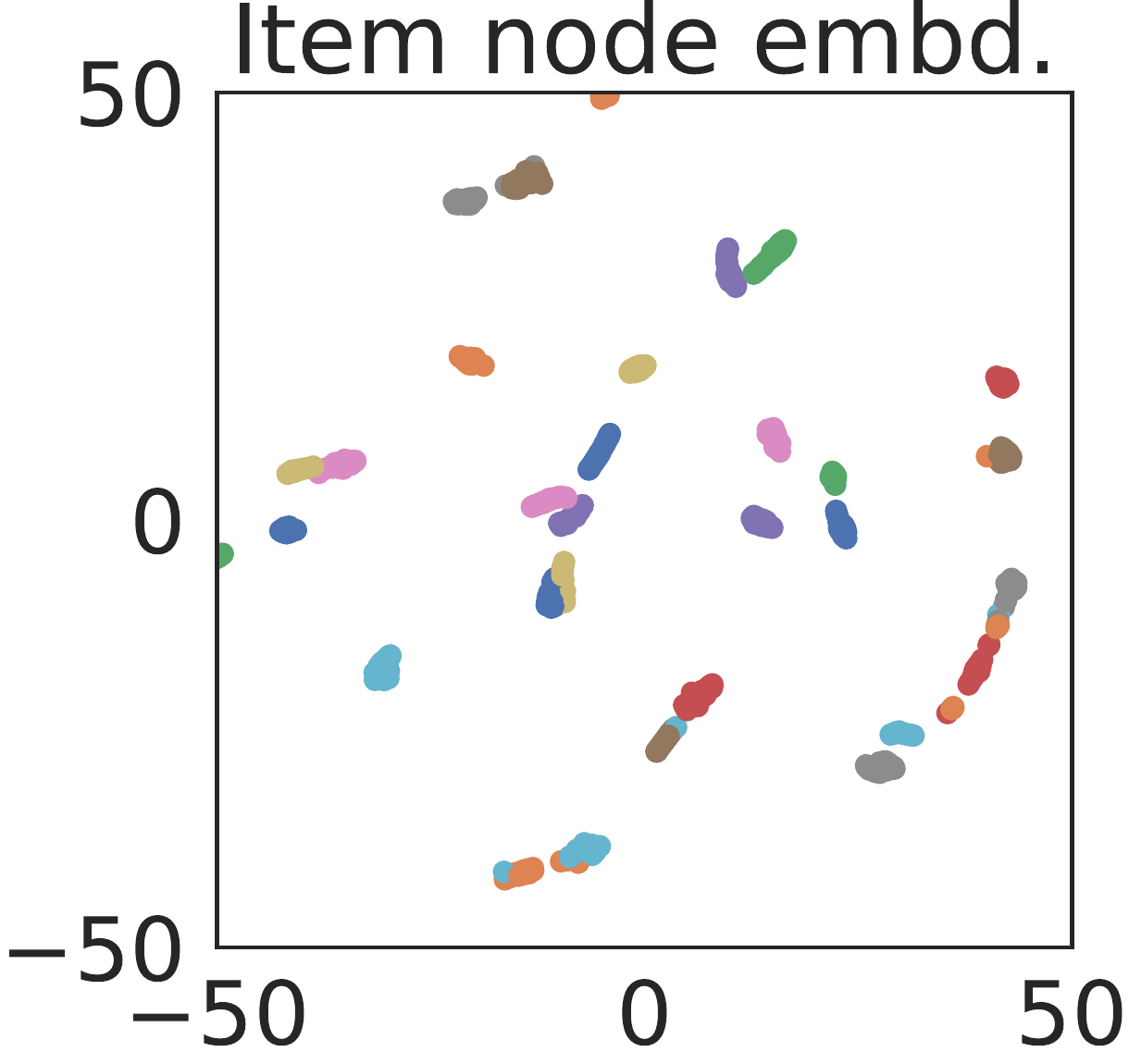}
    \caption{}
    \label{subfig:synthetic_item_node_emb}
  \end{subfigure}
  \begin{subfigure}[b]{0.195\textwidth}
    \centering
    \includegraphics[width=\textwidth]{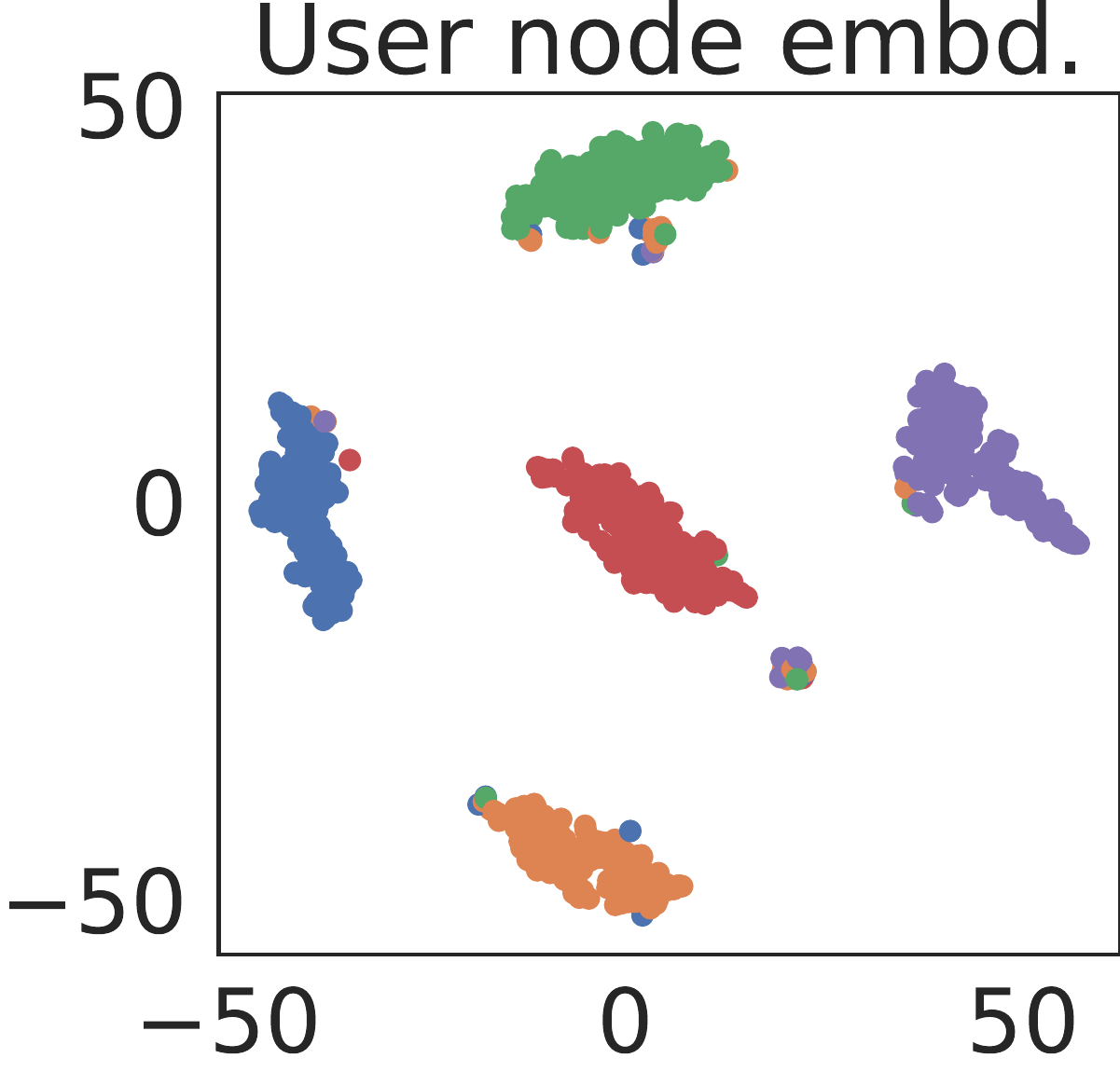}
    \caption{}
    \label{subfig:synthetic_user_node_emb}
  \end{subfigure}
  \begin{subfigure}[b]{0.195\textwidth}
    \centering
    \includegraphics[width=\textwidth]{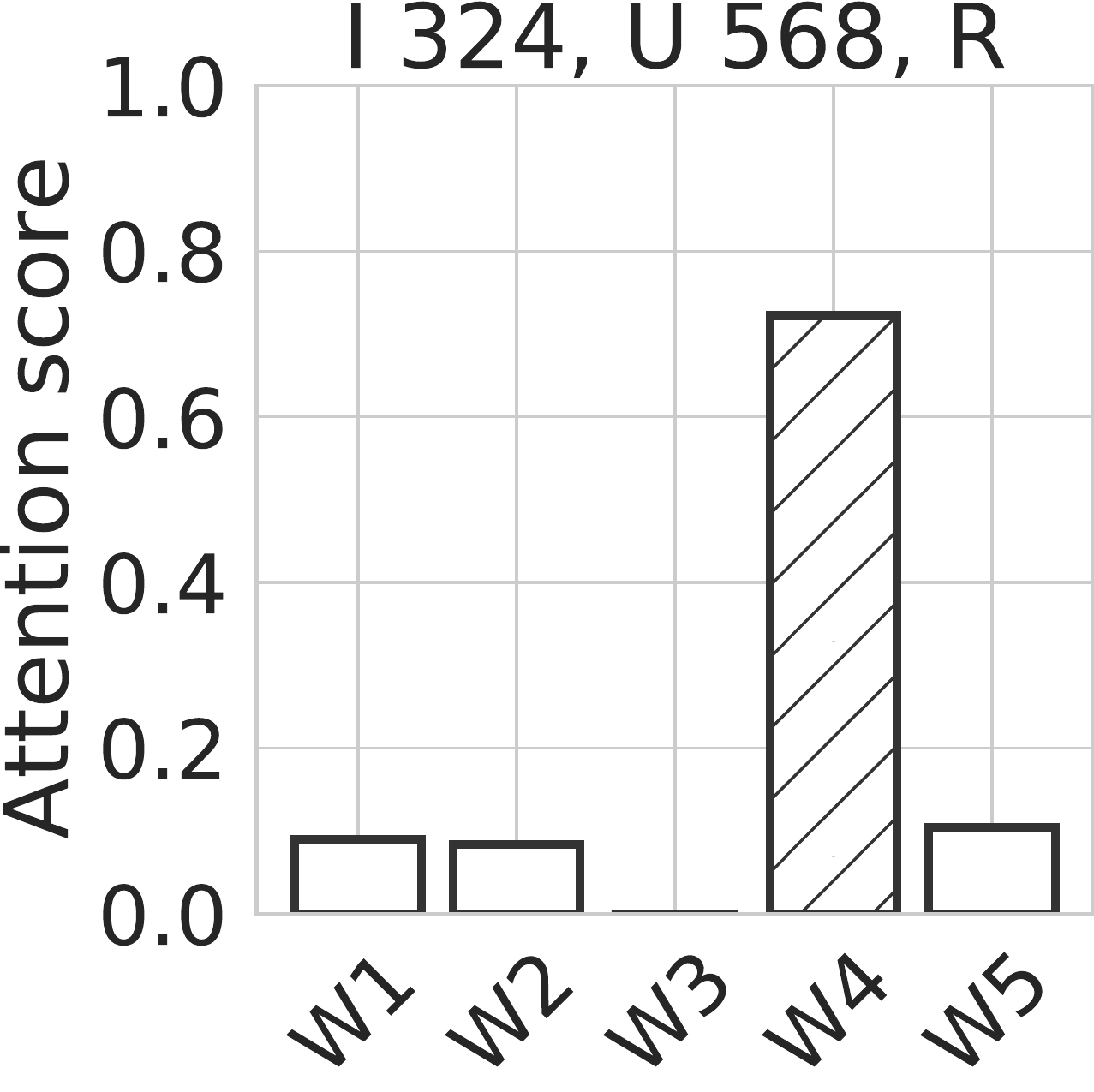}
    \caption{}
    \label{subfig:synthetic_right}
  \end{subfigure}
  \begin{subfigure}[b]{0.195\textwidth}
    \centering
    \includegraphics[width=\textwidth]{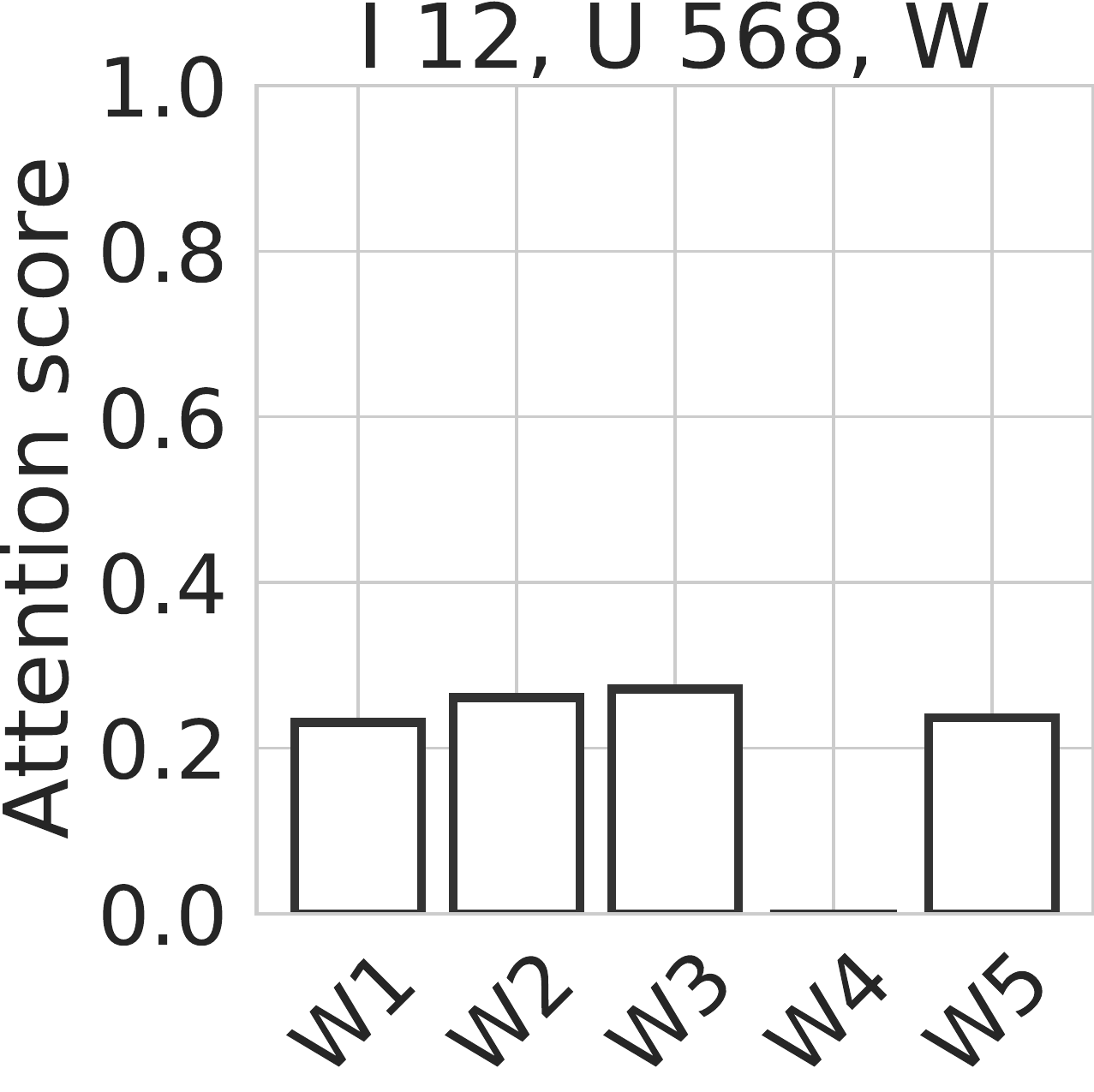}
    \caption{}
    \label{subfig:synthetic_wrong}
  \end{subfigure}
  \caption{
      Synthetic experiment results. 
      (a): Test accuracies when using item content information for node initialization (N.I.), for computing content attentions (W.A.), and when the the edge labels are given (L$*$).
      (b), (c): $t$-SNE plots of learned item and user node embeddings using \modelName.
      Each dot represents a single node and is colored by its word distribution (item) and word-attentiveness (user).
      (d), (e): Computed word-attention scores of a user with correct and incorrect answers. 
  }
  \label{fig:synthetic}
\end{figure*}
\subsection{Synthetic experiments}
First, we create a synthetic dataset to validate our model design.
We create a random bipartite graph with item and user nodes. Each item-node is associated with a number of ``words'' as its content.
We use a ``vocabulary'' of 5 words and each item node contain each of the 5 words with 50\% probability.
Each user-node is assigned a single \emph{focus word} that indicates the word the user ``likes''.
Finally, the value of an edge between a user and an item is deterministically set to 1 if the item contains the user's ``focus word'', and 0 otherwise.
Throughout experiments, the word-content of items is provided as an input to \modelName, but the user focus word is latent.
We are interested in understanding if \modelName can predict the correct edge labels between users and items, which can easily be achieved using content.
Furthermore, by inspecting attention scores, we analyze \modelName's ability to learn to focus on the user's focus word within each item node, if it is present.

Fig.~\ref{subfig:synthetic_test_acc} shows the test accuracy of GCN with item node initialization using word vectors (left, blue bar), \modelName (middle, red bar), and GCN with edge embeddings initialized with edge labels (right, yellow bar).
Unlike the last model, the first two models do not use the ground truth edge labels during training.
\modelName is the only one that achieves near perfect test accuracy.

Fig.~\ref{subfig:synthetic_item_node_emb} and Fig.~\ref{subfig:synthetic_user_node_emb} illustrate the $t$-SNE~\citep{van2008visualizing} visualization of computed user-node and item-node embeddings for \modelName.
Item and user nodes are colored by their associated content-word distributions and word-attentiveness, showing that the node embeddings can discriminate nodes by their attributes.
Fig.~\ref{subfig:synthetic_right} and Fig.~\ref{subfig:synthetic_wrong} display the computed attention scores between user-item pairs for two sample pairs.
When the word that the user ``likes'' is included in the item's associated words, \modelName correctly targets that word by assigning high attention score~(Fig.~\ref{subfig:synthetic_right}).
When word that user ``likes'' is absent, the attention distribution over the content-words of an item-node becomes much more uniform.

\subsection{Evaluation on real-world data}
\paragraph{Datasets}
We evaluate \modelName on two real-world datasets
that record different user-item interactions (Tbl.~\ref{tab:data_statistics}).
The Goodreads dataset~\citep{goodreads2020} from the Goodreads website 
contains users and books. The content of each book-node is its natural language description. 
The dataset includes a 1 to 5 integer rating between some books and users.
The Eedi dataset~\citep{wang2020diagnostic} contains anonymized student and question identities with the student responses to some questions. The content of each question-node is the text of the question. %
Edge labels are binary: one and zero for correct and incorrect answers.

For both datasets, to encode the content within nodes in \modelName, we use a pre-trained transformer encoder model~\citep{devlin2019bert} as $\mathbf{F}$.
We use a \emph{truncation threshold} $\mathit{T}$ so that we ignore words that appear after $\mathit{T}$ for any $\mathbf{D}_m$ with $n(m) > \mathit{T}$.
In our experiments, the parameters of the GNN
of \modelName and the prediction multi-layer perceptron (MLP) (Eq.~\ref{eq:recommendation prediction}) are learned jointly during training but we do \emph{not} fine-tune the parameters of $\mathbf{F}$.
We set $T = 64$ for both Goodreads and Eedi.
The appendix compares test performance with respect to varying $T$ for Goodreads and Eedi and additional information about the dataset and the pre-processing steps.

\begin{table*}[t]\centering\footnotesize
\caption{Dataset statistics: $\vert \mathcal{D}\vert$: Vocabulary size, $\vert \overline{\mathbf{D}} \vert$: avg num of words per item, Density: graph density, \#L: num of labels.}
\label{tab:data_statistics}

\begin{tabular}{@{}lrrrrrrrrr@{}}
\toprule
 & \multicolumn{2}{c}{Nodes}                                  & \multicolumn{3}{c}{Edges}                                                                  & \multicolumn{2}{c}{Contents}                                          & \multicolumn{1}{c}{\multirow{2}{*}{Density}} 
       & \multicolumn{1}{c}{\multirow{2}{*}{\# L}}\\ \cmidrule(lr){2-3} \cmidrule(lr){4-6} \cmidrule(lr){7-8}
                  & \multicolumn{1}{c}{\# Users} & \multicolumn{1}{c}{\#Items} & \multicolumn{1}{c}{\# Edges} & \multicolumn{1}{c}{/ user} & \multicolumn{1}{c}{/ item} & \multicolumn{1}{c}{$\vert \mathcal{D}\vert$} & \multicolumn{1}{c}{$\vert \overline{\mathbf{D}} \vert$} & \multicolumn{1}{c}{}   & \multicolumn{1}{c}{}                        \\ \midrule
\textbf{Synthetic}         & 1,000                        & 1,000                       & 100,000                    & 100                       & 100                       & 5                               & 2.50                                  & 0.100 & 2                                              \\ 
\textbf{Eedi}              & 35,073                        & 22,931                         & 991,740                    & 28                       & 43                     & 21,072                           & 20.02                               & 0.001 & 2                                              \\ 
\textbf{Goodreads}         & 2,243                        & 2,452                       & 114,839                      & 51                        & 47                        & 35,111                          & 132.32                              & 0.021 & 5                                              \\ \bottomrule
\end{tabular}
\end{table*}

\begin{table*}[t!]
\centering
\caption{
    Average test RMSE (Goodreads, lower is better) and test accuracy, AUROC, AUPR (Eedi, higher is better) results over 5 independent runs followed by one standard error. 
    Best results are highlighted in bold, and the second-best results are underlined. 
    $*$ and $**$ signify $p$-values less than 0.05 and 0.001 respectively from independent $t$-tests with the second-best results.
}
\label{tab:performance}\footnotesize
\begin{tabular}{@{}lccrrr@{}}
\toprule
Model &
Content &
  \multicolumn{1}{c}{\qquad \textbf{Goodreads \qquad\qquad}} &
  \multicolumn{3}{c}{\textbf{Eedi}} \\ \cmidrule(l){3-3} \cmidrule(l){4-6}
 & &
  \multicolumn{1}{c}{RMSE ($\downarrow$)} &
  \multicolumn{1}{c}{Accuracy ($\uparrow$)} &
  \multicolumn{1}{c}{AUROC ($\uparrow$)} &
  \multicolumn{1}{c}{AUPR ($\uparrow$)} \\ \midrule
DMF~\citep{xue2017deep}&\xmark &
  0.921${}_{\pm0.001}$ &
  0.738${}_{\pm0.002}$ &
  0.653${}_{\pm0.003}$ &
  0.828${}_{\pm0.002}$ \\
PVAE~\citep{ma2019eddi} &\xmark &
  0.894${}_{\pm0.001}$ &
  0.746${}_{\pm0.001}$ &
  0.682${}_{\pm0.000}$ &
  0.834${}_{\pm0.000}$ \\\midrule
GC-MC~\citep{berg2017graph} &\xmark &
  0.916${}_{\pm0.002}$ &
  0.735${}_{\pm0.001}$ &
  0.672${}_{\pm0.002}$ &
  0.819${}_{\pm0.001}$ \\
GCN~\citep{kipf2016semi}& \xmark&
  0.893${}_{\pm0.001}$ &
  0.746${}_{\pm0.002}$ &
  0.680${}_{\pm0.001}$ &
  0.830${}_{\pm0.001}$ \\
GraphSAGE~\citep{hamilton2017inductive} &\xmark &
  0.898${}_{\pm0.003}$ &
  0.742${}_{\pm0.003}$ &
  0.665${}_{\pm0.002}$ &
  0.838${}_{\pm0.003}$ \\
GRAPE~\citep{you2020handling} &\xmark &
  0.894${}_{\pm0.001}$ &
  0.746${}_{\pm0.001}$ &
  0.672${}_{\pm0.001}$ &
  0.824${}_{\pm0.001}$ \\
GAT~\citep{velivckovic2017graph} &\xmark &
  0.893${}_{\pm0.002}$ &
  0.745${}_{\pm0.000}$ &
  0.684${}_{\pm0.001}$ &
  0.832${}_{\pm0.001}$ \\
GIN-$\epsilon$~\citep{xu2018powerful}&\xmark &
  0.892${}_{\pm0.000}$ &
  0.747${}_{\pm0.001}$ &
  0.689${}_{\pm0.002}$ &
  0.838${}_{\pm0.001}$ \\
JK-LSTM~\citep{xu2018representation}&\xmark &
  0.895${}_{\pm0.000}$ &
  0.746${}_{\pm0.001}$ &
  0.685${}_{\pm0.002}$ &
  0.840${}_{\pm0.002}$ \\ \midrule
GCN with WordNodes & \cmark &
  0.886${}_{\pm0.002}$ &
  0.751${}_{\pm0.002}$ &
  0.710${}_{\pm0.002}$ &
  0.839${}_{\pm0.003}$ \\
GCN Init: BoW & \cmark &
  0.891${}_{\pm0.001}$ &
  0.748${}_{\pm0.001}$ &
  0.710${}_{\pm0.001}$ &
  0.836${}_{\pm0.000}$ \\
GCN Init: NeuralBoW & \cmark &
  0.886${}_{\pm0.001}$ &
  0.751${}_{\pm0.000}$ &
  0.706${}_{\pm0.001}$ &
  0.848${}_{\pm0.001}$ \\
GCN Init: BERT CLS &\cmark &
  0.889${}_{\pm0.001}$ &
  0.748${}_{\pm0.001}$ &
  0.706${}_{\pm0.001}$ &
  0.841${}_{\pm0.001}$ \\
GCN Init: BERT Avg. &\cmark &
  0.887${}_{\pm0.001}$ &
  0.750${}_{\pm0.001}$ &
  0.708${}_{\pm0.001}$ &
  0.848${}_{\pm0.001}$ \\
GCN Init: SBERT &\cmark &
  0.890${}_{\pm0.000}$ &
  0.752${}_{\pm0.002}$ &
  0.708${}_{\pm0.002}$ &
  0.848${}_{\pm0.001}$ \\ 
GAT Init: NeuralBoW & \cmark &
  0.888${}_{\pm0.001}$ &
  0.747${}_{\pm0.001}$ &
  0.707${}_{\pm0.002}$ &
  0.841${}_{\pm0.001}$ \\ 
GAT Init: BERT Avg. &\cmark &
  0.887${}_{\pm0.001}$ &
  0.749${}_{\pm0.001}$ &
  0.709${}_{\pm0.001}$ &
  0.844${}_{\pm0.002}$ \\
GAT Init: SBERT &\cmark &
  \underline{0.884}${}_{\pm0.000}$ &
  0.752${}_{\pm0.001}$ &
  0.710${}_{\pm0.001}$ &
  \underline{0.849}${}_{\pm0.001}$ \\
GRAPE Init: NeuralBoW & \cmark &
  0.890${}_{\pm0.001}$ &
  0.748${}_{\pm0.001}$ &
  0.704${}_{\pm0.001}$ &
  0.842${}_{\pm0.001}$ \\ 
GRAPE Init: BERT Avg. &\cmark &
  0.889${}_{\pm0.001}$ &
  0.749${}_{\pm0.000}$ &
  0.704${}_{\pm0.001}$ &
  0.838${}_{\pm0.001}$ \\
GRAPE Init: SBERT &\cmark &
  0.892${}_{\pm0.001}$ &
  0.750${}_{\pm0.002}$ &
  0.711${}_{\pm0.002}$ &
  0.846${}_{\pm0.003}$ \\
GIN Init: NeuralBoW & \cmark &
  0.890${}_{\pm0.001}$ &
  0.748${}_{\pm0.002}$ &
  0.710${}_{\pm0.001}$ &
  0.843${}_{\pm0.002}$ \\ 
GIN Init: BERT Avg. &\cmark &
  0.888${}_{\pm0.001}$ &
  0.752${}_{\pm0.001}$ &
  \underline{0.711}${}_{\pm0.001}$ &
  0.836${}_{\pm0.002}$ \\
GIN Init: SBERT &\cmark &
  0.885${}_{\pm0.000}$ &
  \underline{0.752}${}_{\pm0.001}$ &
  0.711${}_{\pm0.000}$ &
  0.845${}_{\pm0.001}$ \\
JK Init: NeuralBoW & \cmark &
  0.889${}_{\pm0.001}$ &
  0.747${}_{\pm0.002}$ &
  0.709${}_{\pm0.001}$ &
  0.842${}_{\pm0.000}$ \\ 
JK Init: BERT Avg. &\cmark &
  0.889${}_{\pm0.001}$ &
  0.749${}_{\pm0.002}$ &
  0.708${}_{\pm0.001}$ &
  0.838${}_{\pm0.001}$ \\
JK Init: SBERT &\cmark &
  0.886${}_{\pm0.000}$ &
  0.751${}_{\pm0.001}$ &
  0.710${}_{\pm0.001}$ &
  0.839${}_{\pm0.002}$ \\ \midrule
\modelName: Concat &\cmark &
  \textbf{0.873}${}^{**}_{\pm0.000}$ &
  \textbf{0.761}${}^{**}_{\pm0.001}$ &
  \textbf{0.720}${}^{**}_{\pm0.001}$ &
  \textbf{0.891}${}^{**}_{\pm0.000}$ \\
\modelName: Dot-product &\cmark &
  \textbf{0.872}${}^{**}_{\pm0.000}$ &
  \textbf{0.760}${}^{**}_{\pm0.001}$ &
  \textbf{0.721}${}^{**}_{\pm0.001}$ &
  \textbf{0.888}${}^{**}_{\pm0.001}$ \\ \bottomrule
\end{tabular}
\end{table*}
\paragraph{Baselines}
We compare \modelName with 9 widely used missing value imputation models, discussed in related work.
Deep Matrix Factorization (DMF) and Partial Variational Autoencoder (PVAE) are non-GNN matrix completion models.
Graph Convolutional Network (GCN), GraphSAGE, Graph Attention Network, Jumping knowledge network, Graph isomorphism network are GNN-based models not specifically designed for recommender systems.
We compare \modelName with these baselines by using a read-out MLP that accepts the concatenation of user and item node embeddings and makes a
prediction for the pair, as done in Eq.~\ref{eq:recommendation prediction}.
We also compare to graph convolutional matrix completion (GC-MC) and GRAPE~\citep{you2020handling} that are GNN-based models for matrix completion.

None of the previous models consider content. Thus, we consider 6 GNN node embedding initialization configurations that use content information: 

\begin{description}
    \item \emph{GNNs with WordNodes}. We create special ``word nodes'' for every word and connect them to a node $v_m \in \mathcal{V}_M$ if the word is contained in the item's content.
    This baseline allows message passing word-specific information but word ordering within item content is ignored.
    We retrieve words by stemming~\citep{porter1980algorithm}, filtering non-alphanumeric words, and removing words with frequency of less than 2.
    %
    %
    \item \emph{GNN Node Init: BoW} is a standard GCN with the node embeddings initialized to a multi-hot bag-of-words of the content. Words are tokenized and stemmed as above.
    %
    %
    \item \emph{GNN Node Init: NeuralBoW} uses a pre-trained word2vec model~\citep{mikolov2013distributed}, implemented in Gensim~\citep{rehurek_lrec}. Words are encoded in fixed 300-dimensional vector representations and average pooling is used for node initialization.
    %
    \item \emph{GNN Node Init: BERT CLS} uses pre-trained, cased BERT-base~\citep{devlin2019bert}, impemented in HuggingFace~\citep{wolf2019huggingface}. Words are encoded in 768-dimensional vector representations.
    and the representation of \texttt{[CLS]} is used to initalize node embeddings.
    %
    %
    \item \emph{GNN Node Init: BERT Avg} uses identical settings as above, but instead of the \texttt{[CLS]} token, we use average pooling over all the output vector representations.
    %
    %
    \item \emph{GNN Node Init: SBERT} uses the 768-dimensional vector representations encoding of the whole document with the SBERT model of \citet{reimers2019sentence}.
    %
    %
%
\end{description}
In all cases, the content vector is mapped to the initial node embeddings using a learned linear layer.
\begin{figure}[t]
    \centering
    \captionsetup{type=figure}
    \includegraphics[width=0.8\linewidth]{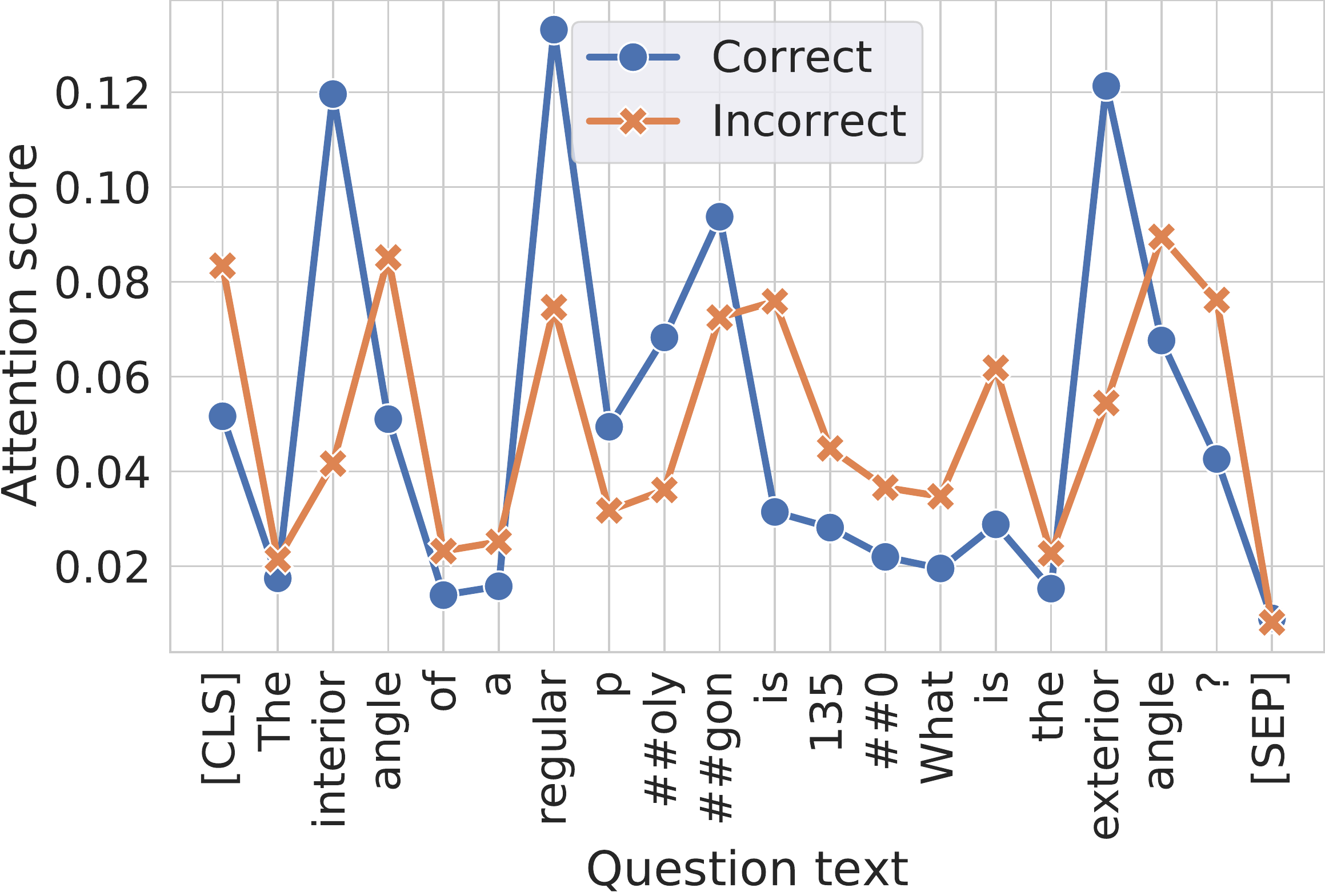}
    \caption{Example content attention distributions of students with correct and incorrect answers.}
    \label{fig:word_attention}
\end{figure}
    
%
\begin{table*}[t!]
\centering
\caption{
    Mean $\pm$ one standard error of rating prediction over 5 runs (Goodreads, test RMSE) and response prediction (Eedi, test accuracy) results of all users (All), users with node degree greater than 10 ($D > 10$), and users with degrees less than or equal to 10 ($D \leq 10$).
    $*$ and $**$ signify $p$-values less than 0.05 and 0.001 respectively from paired $t$-tests.
}
\label{tab:sparsity}
\footnotesize
\begin{tabular}{@{}lrrrrrr@{}}
\toprule
\multicolumn{1}{c}{\multirow{2}{*}{}} & \multicolumn{3}{c}{\textbf{Goodreads} - RMSE ($\downarrow$)}                                                                        & \multicolumn{3}{c}{\textbf{Eedi} - Accuracy ($\uparrow$)}                                                                             \\ \cmidrule(l){2-4}\cmidrule(l){5-7} 
\multicolumn{1}{c}{}                  & \multicolumn{1}{c}{All} & \multicolumn{1}{c}{D $>$ 10} & \multicolumn{1}{c}{D $\leq$ 10} & \multicolumn{1}{c}{All} & \multicolumn{1}{c}{D $>$ 10} & \multicolumn{1}{c}{D $\leq$ 10} \\ \midrule
GCN                                   & ${0.792}^{}_{\pm 0.007}$             & ${0.796}^{}_{\pm 0.008}$                           & ${0.752}^{}_{\pm 0.03}$                         & ${0.746}^{}_{\pm 0.002}$             &${0.751}^{}_{\pm 0.002}$                           & ${0.720}^{}_{\pm 0.008}$                        \\
GCN \small{N.I.: SBERT}                     & ${0.786}^{}_{\pm 0.006}$             & ${0.789}^{}_{\pm 0.007}$                           & ${0.728}^{}_{\pm 0.03}$                         & ${0.753}^{}_{\pm 0.003}$             & ${0.757}^{}_{\pm 0.002}$                           & ${0.727}^{}_{\pm 0.008}$                        \\
\textbf{\modelName (DP)}                    & ${\bm{0.781}}^{*}_{\pm 0.006}$    & ${\bm{0.786}}^{}_{\pm 0.007}$                  & ${\bm{0.685}}^{**}_{\pm 0.03}$                & ${\bm{0.758}}^{*}_{\pm 0.002}$    & ${\bm{0.760}}^{*}_{\pm 0.003}$                  & ${\bm{0.750}}^{**}_{\pm 0.010}$               \\ \bottomrule
\end{tabular}

\end{table*}
\paragraph{Missing value imputation}
Tbl.~\ref{tab:performance} compares the missing value imputation performance on the Goodreads and Eedi datasets over 5 runs.
We report root mean square error (RMSE) for Goodreads and accuracy, area under the receiver operating characteristic (AUROC), and area under the precision-recall curve (AUPR) for Eedi.
Overall, we observe improved performance when node content is used.
Compared to the baseline models with content, both \modelName constructions (concatenation and dot-product) outperform baselines on both datasets with statistical significance. 

Fig.~\ref{fig:word_attention} shows the content attention distributions of user (student) - item (question) pairs in Eedi dataset for a particular question. 
The blue circle line shows the average attention scores of students who got the question right, and the orange cross line shows that of students with incorrect answers.
We observe \emph{user-item-specific} attention scores assigned; the student with right answer has high attention scores for tokens \texttt{interior}, \texttt{regular}, and \texttt{exterior}, while the student with wrong answer attends more to tokens \texttt{angle}, \texttt{angle}, and \texttt{[CLS]}.
This is in contrast to baselines that exploit the content information of items in a way that does not explicitly distinguish users during the message passing.  

%
\paragraph{Sparsity analysis}
\begin{figure}[tb]\centering
    \includegraphics[width=0.6\columnwidth]{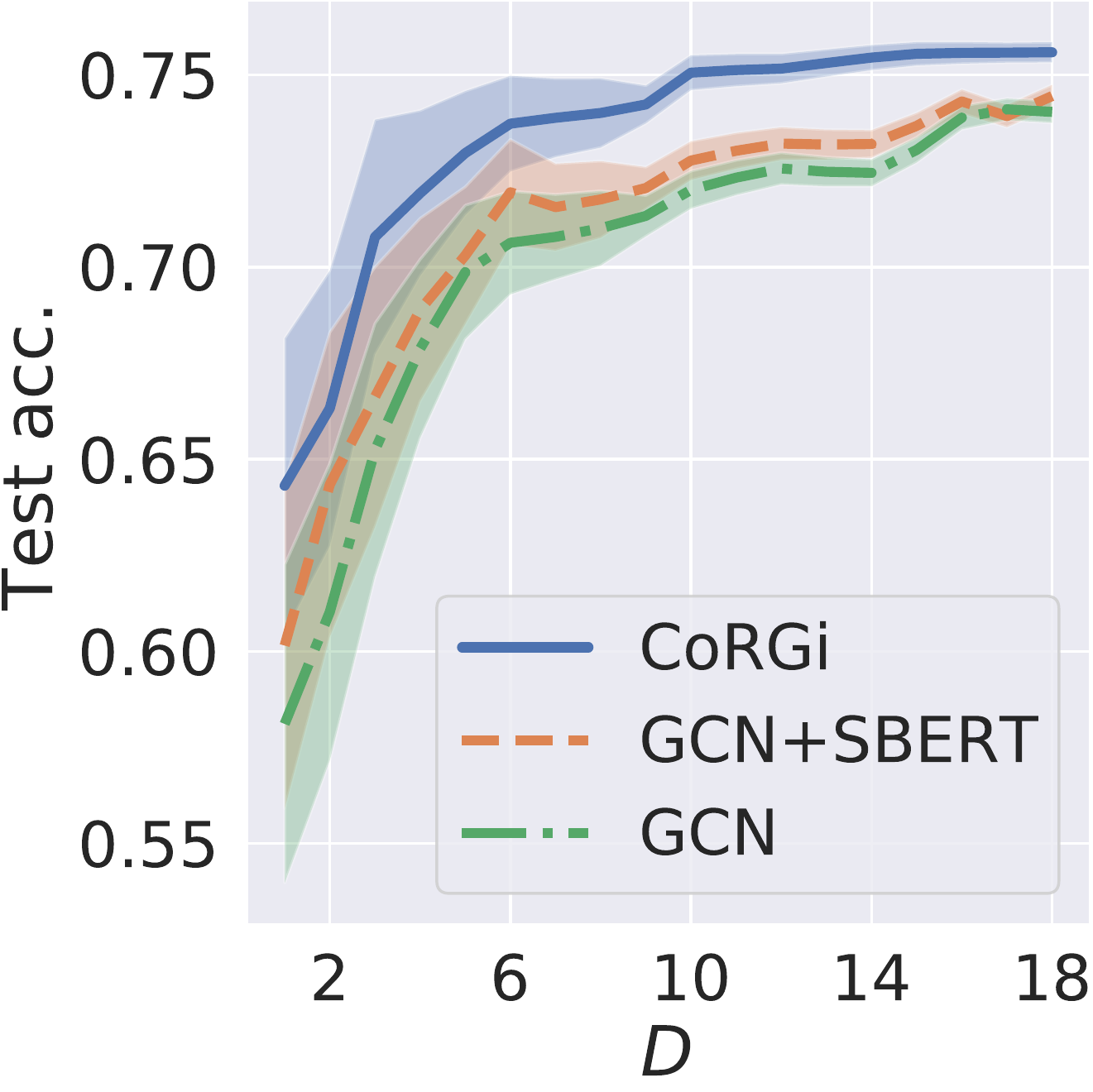}
    \caption{Test accuracy vs. user node degree for Eedi.}
    \label{fig:sparsity}
\end{figure} 

Finally, we test the hypothesis that content is particularly useful in sparse --- less connected --- regions of the recommendation.
Tbl.~\ref{tab:sparsity} shows rating and response prediction performance on Goodreads and Eedi vs. the user node degree ($D$),
, i.e., the number of questions answered or books rated.
\modelName outperforms the baselines
with $p<0.05$ from paired $t$-tests with respect to GCN Node Init: SBERT. 
On both datasets, the predictive performance between \modelName and the comparison models is relatively comparable for users that have interacted with more items ($D > 10$), although \modelName still outperforms them.
On the contrary, the difference in performance between \modelName and the comparison models becomes more significant for users connected with $D \leq 10$, showing the effectiveness of \modelName.
Fig.~\ref{fig:sparsity} shows test accuracy on Eedi with varying degrees, showing an increasing gap between \modelName and baselines with smaller $D$s.
\begin{table}[t]
\vspace{-2mm}
\caption{Win rates (column vs. row) in inductive setting for two held-out rates (0.5 and 0.99).}
\vspace{-3mm}
\centering
\begin{tabular}{@{}crr@{}}
\toprule
\multicolumn{1}{l}{} & \multicolumn{1}{l}{GRAPE} & \multicolumn{1}{l}{Init: SBERT}  \\ \midrule

Init: SBERT          & 0.514 / 0.610                     & \multicolumn{1}{c}{-}                                      \\
\textbf{\modelName}                & 0.532 / 0.837                     & 0.526 / 0.695                                \\ \bottomrule
\end{tabular}
\label{tbl:inductive1}
\end{table}
%

Tbl.~\ref{tbl:inductive1} show the relative predictive performance of \modelName over the comparison models in inductive settings, i.e., prediction on users that were not seen during training but newly introduced during inference, in the Eedi dataset. We evaluate win-rate, the rate at which a model’s prediction is correct and the other model’s prediction is wrong. We find that when the train graph becomes sparse with larger number of unseen nodes during training, \modelName's relative predictive performance compared to the baseline models improves compared to the dense setting.

%% file: 5_conclusion.tex
\section{Conclusion}
\label{section:conclusion}
We presented \modelName, a message-passing GNN that tightly integrates node content using attention. %
Using node content --- such as text --- allows us to capture rich
information within the modeled domain while exploiting the
structured form of the data. This is
particularly evident in sparse regions of graphs.
Future work may further investigate how non-text modalities can be captured in content-rich graphs across a range of applications beyond edge value prediction.

\clearpage

%% file: appendix.tex
\appendix

\section{Summary of key notations used in the paper}
\label{appendix:notations}

\renewcommand{\arraystretch}{1.5}

\begin{table*}[h]
\centering
\caption{Key notations used in the paper.}
\label{tab:notations}
\begin{tabular}{@{}lll@{}}
\toprule
 &
  \multicolumn{1}{c}{Symbols} &
  \multicolumn{1}{c}{Description} \\ \midrule
\multirow{4}{*}{\begin{tabular}[c]{@{}l@{}}\textbf{Graph sets}\\ \& \textbf{elements}\end{tabular}} &
  $\mathcal{V}$ &
  The set of all nodes in the graph \\
 &
  $\mathcal{V}_{C}, \mathcal{V}_M, \mathcal{V}_U \subset \mathcal{V}$ &
  The sets of content, item, and user nodes \\
 &
  $\mathcal{E}$ &
  The set of all edges in the graph \\
 &
  $\mathcal{N}(i)$ &
  Neighborhood function for node $v_i$ \\ \midrule
\multirow{8}{*}{\begin{tabular}[c]{@{}l@{}}\textbf{\modelName}\\ \textbf{variables}\end{tabular}} &
  $\mathbf{h}_i^{(0)}$ &
  Input feature of node $v_i$ of size $C^{(l, h)}$ \\
 &
  $\mathbf{e}_{ij}^{(0)}$ &
  Input feature of edge $e_{ij}$ of size $C^{(l, e)}$ \\
 &
  $\mathbf{h}_i^{(l)}$ &
  Node embedding of $v_i$ at $l^{th}$ layer \\
 &
  $\mathbf{e}_{ij}^{(l)}$ &
  Edge embedding between $v_i$ and $v_j$ at $l^{th}$ layer \\
 &
  $\mathbf{e}_{ij}^{(l)\prime}$ &
  Edge embedding before content update \\
 &
  $\mathbf{e}_{ij, \text{CA}}^{(l)}$ &
  Edge embedding from content-attention \\
 &
  $c_{ik}^{(l)}$ &
  Attention coefficient between $v_i$ and content $k$ \\
 &
  $\alpha_{ik}^{(l)}$ &
  Attention probability from $c_{ik}^{(l)}$ after \textsc{Softmax} \\ \midrule
\multirow{4}{*}{\begin{tabular}[c]{@{}l@{}}\textbf{Content-related}\\ \textbf{notations}\end{tabular}} &
  $n(i)$ &
  The number of content vectors associated to $v_i$ \\
 &
  $\mathbf{Z}_{i} = \{ \mathbf{z}_k^{(\mathit{i})}  \}_{1}^{n(i)} \subset \mathbb{R}^D  $ &
  A set of content vectors associated to $v_i$ \\
 &
  $\mathbf{D}_i = [w_1^{(i)}, \cdots, w_{n(i)}^{(i)}]$ &
  A sequence of words associated to $v_i$ \\
 &
  $\mathbf{F}$ &
  A sequence encoder that projects $\mathbf{D}_i$ to $\mathbf{Z}_i$ \\ \bottomrule
\end{tabular}
\end{table*}

We summarize the key notations used throughout the paper.
We group notations in three groups: 
(a) notations on graph sets and the corresponding elements.
(b) variables and parameters used to describe the forward pass of \modelName.
(c) notations used to describe the content information associated to content or item nodes.
In addition to this, we have the trainable weights explained during the message passing of \modelName:
$\mathbf{P}^{(l)}$ and $\mathbf{Q}^{(l)}$ for updating node embeddings,
$\mathbf{W}_U^{(l)}$, $\mathbf{W}_M^{(l)}$, and $\mathbf{p}^{(l)}$ for computing attention coefficients,
and $\mathbf{w}_{\text{out}}$ and $b$ for the prediction MLP.

\clearpage

\section{Additional experiments}
\renewcommand{\arraystretch}{1.2}
\subsection{Truncation threshold and test performance}
\label{appendix:truncation}
\begin{figure}[ht]
    \centering
    \includegraphics[width=0.8\linewidth]{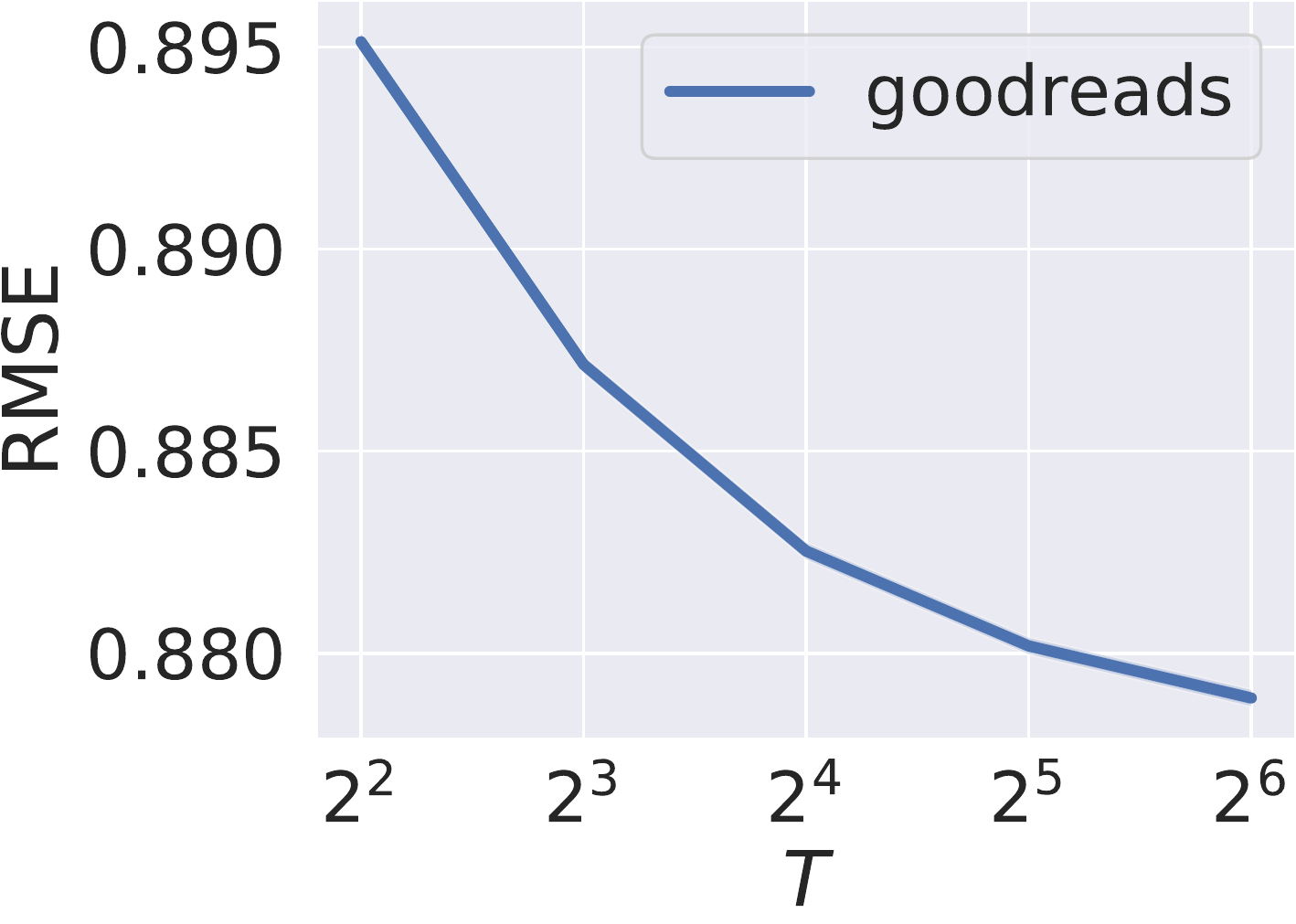}
    \caption{Truncation size and test accuracy for the Goodreads dataset.  Note semi-log-$x$ and starting point for $y$-axis.}
    \label{fig:appendix_truncation_eedi}
\end{figure}
\begin{figure}[ht]
    \centering
    \includegraphics[width=0.8\linewidth]{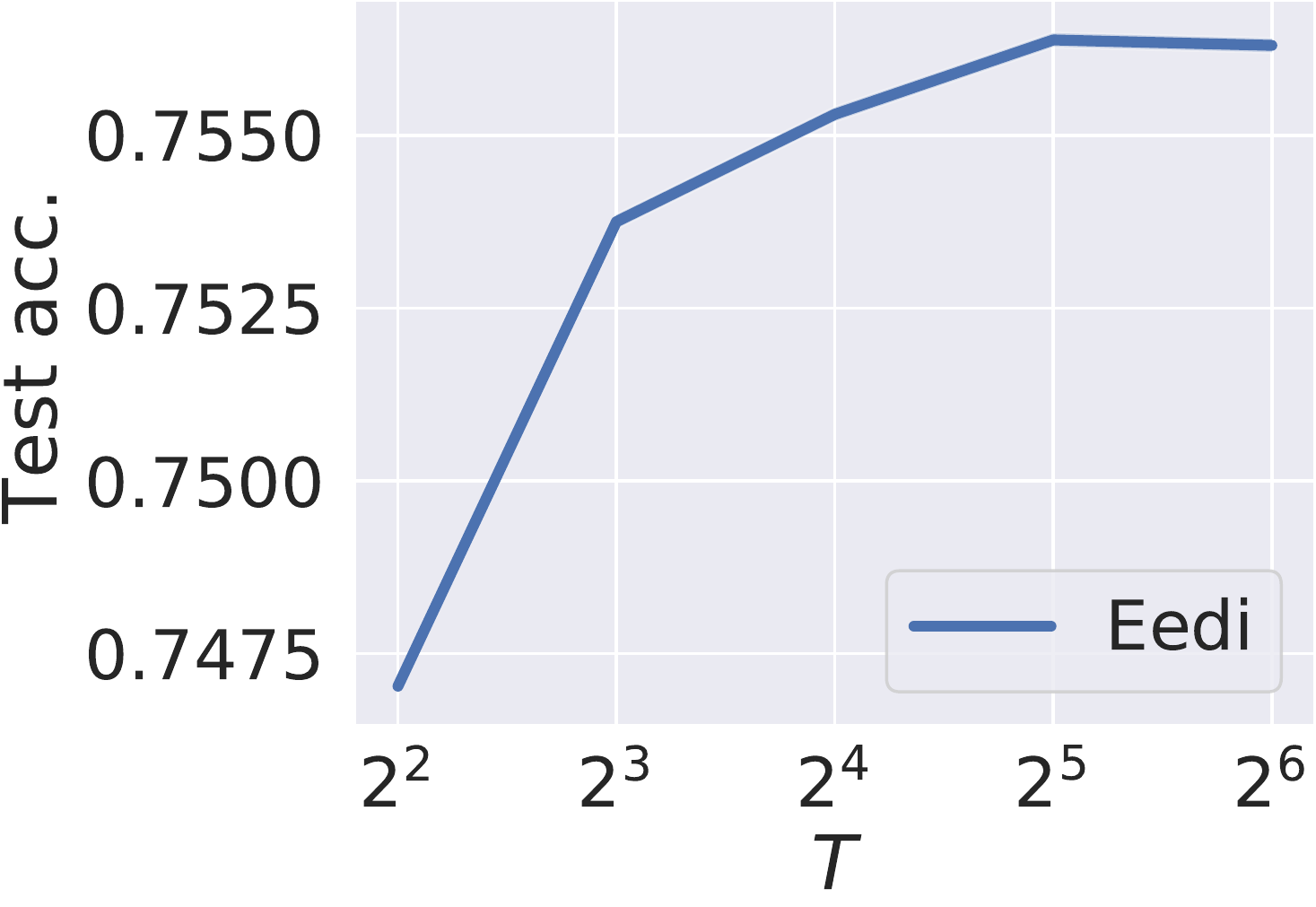}
    \caption{Truncation size and RMSE for the Eedi dataset. Note semi-log-$x$ and starting point for $y$-axis.}
    \label{fig:appendix_truncation_goodreads}
\end{figure}
We test the affect of the truncation threshold $T$ on the test performance.
In sequential content encoders $\mathbf{F}$ such as the transformer encodes $\mathbf{D_m}$, i.e., the contents of an item node $v_m$ of size $n(m)$ into a set of content vector representations $\mathbf{Z}_m$.
During encoding, if $n(m)$ is greater than the truncation threshold $T$, we only the first $T$ words only, i.e., $\vert \mathbf{Z}_m \vert = \min (n(m), T)$.
Setting $T$ to a high value enables \modelName to fully exploit the content information, at a trade-off that makes the models slow to train with larger memory requirement.

Fig.~\ref{fig:appendix_truncation_eedi} and~\ref{fig:appendix_truncation_goodreads} show the test performance with respect to varying values of $T$ on the Goodreads and Eedi datasets.
For both datasets, increasing $T$ results in higher test performance ($T$ greater than 64 results in the memory error on our computing infrastructure).
The average number of words on Eedi questions 20.02 (Tbl.~\ref{tab:data_statistics}), and the test performance converges at $T=32$.
On the other hand, the average number of words on Goodreads book descriptions is 132.32, and we observe that the test performance does not fully converge at $T=64$.

\subsection{Performance comparison on caching trick}
In Sec.~\ref{subsection:complexity}, we introduce a caching trick that allows the reduction in training time and memory requirement.
Specifically, the caching trick is realized by creating a cache for $\mathbf{e}_{ij, \text{CA}}^{(l)}$ with zero initializations.
We then update at the final layer $L$ only, by computing $\mathbf{e}_{ij, \text{CA}}^{(L)}$ using Eq.~\ref{eqn:edge_emb} and updating the cache for all $\mathbf{e}_{ij, \text{CA}}^{(l)}$ to the computed $\mathbf{e}_{ij, \text{CA}}^{(L)}$.
Using the caching trick along with the neighbor sampling~\citep{hamilton2017inductive}, the time complexity reduces from
\begin{multline*}
\mathcal{O} \Big (
    \vert \mathcal{V}_U \vert \cdot \mathit{C}^{(l-1, h)} \cdot \mathit{C}^{(l, e)}
    +
    \vert \mathcal{V}_M \vert \cdot \mathit{T} \cdot \mathit{D} \cdot \mathit{C}^{(l, e)} \\
    +
    \vert \mathcal{E}\vert \cdot \mathit{T} \cdot \mathit{C}^{(l, e)}
\Big )
\end{multline*}
to
\begin{multline*}
\mathcal{O} \Big(
    \vert \mathcal{N}(\mathcal{V}_{M}^\prime) \vert \cdot \mathit{C}^{(l-1, h)} \cdot \mathit{C}^{(l, e)}
    +
    \vert \mathcal{V}_M^\prime \vert \cdot \mathit{T} \cdot \mathit{D} \cdot \mathit{C}^{(l, e)} \\
    +
    \vert \mathcal{E}^\prime \vert \cdot \mathit{T} \cdot \mathit{C}^{(l, e)}
\Big),
\end{multline*}

where we sample a subset of nodes $\mathcal{V}^\prime = \big\{ \mathcal{V}_{U}^\prime \cup \mathcal{V}_{M}^\prime \big\}$ for the neighbor sampling and only update $\mathbf{e}_{ij, \text{CA}}$ whose target node $v_j$ is in $\mathcal{V}_{M}^\prime$ and source node $v_i$ is in $\mathcal{N}(\mathcal{V}_{M}^\prime)$.
\renewcommand{\arraystretch}{1.5}
\begin{table}[t!]
\centering
\caption{Performance comparison with and without the caching trick on Goodreads dataset.}
\label{tab:appendix_caching}
\begin{tabular}{@{}lrr@{}}
\toprule
                & \multicolumn{1}{c}{RMSE} & \multicolumn{1}{r}{\begin{tabular}[r]{@{}r@{}}Training time \\ / iteration (sec.)\end{tabular}} \\ \midrule
With caching    & 0.879${}_{\pm 0.000}$                   & 10.41                                                                                           \\
Without caching & 0.879${}_{\pm 0.000}$                    & 41.86                                                                                           \\ \bottomrule
\end{tabular}
\end{table}

Tbl.~\ref{tab:appendix_caching} compares the predictive performance of \modelName with and without the caching trick on Goodreads dataset in terms of RMSE and wall-clock training time.
The predictive performance comparison on Eedi dataset was not feasible due to excessive memory requirement in the absence of the caching trick.
In the absence of the caching trick, the predictive performance remains the same, but the training time is increased more than $4$ times per iteration.
\clearpage

\subsection{Comparison on different combination methods}
\label{appendix:combination_methods}
\begin{table}[t!]
\centering
\caption{
    Comparison of different combination methods for updating the edge embedding $\mathbf{e}_{ij}^{(l)}$ on the Goodreads dataset. 
    We report RMSE (lower the better).
}
\label{tab:appendix_combination_goodreads}
\begin{tabular}{@{}lrr@{}}
\toprule
Combination method &
  \multicolumn{1}{c}{\begin{tabular}[c]{@{}c@{}}\modelName\\ :Concat\end{tabular}} &
  \multicolumn{1}{c}{\begin{tabular}[c]{@{}c@{}}\modelName\\ :Dot-product\end{tabular}} \\ \midrule
$\mathbf{e}_{ij}^{(l)\prime} + \mathbf{e}_{ij, \text{CA}}^{(l)}$ &
  0.879${}_{\pm0.000}$ &
  0.879${}_{\pm0.000}$ \\
$\textsc{Concat} (\mathbf{e}_{ij}^{(0)}, \mathbf{e}_{ij, \text{CA}}^{(l)})$ &
  0.886${}_{\pm0.000}$ &
  0.884${}_{\pm0.001}$ \\ \bottomrule
\end{tabular}
\end{table}
\begin{table}[t!]
\centering
\caption{
    Comparison of different combination methods for updating the edge embedding $\mathbf{e}_{ij}^{(l)}$ on the Eedi dataset. 
    We report test accuracy (higher the better).
}
\label{tab:appendix_combination_eedi}
\begin{tabular}{@{}lrr@{}}
\toprule
Combination method &
  \multicolumn{1}{c}{\begin{tabular}[c]{@{}c@{}}\modelName\\ :Concat\end{tabular}} &
  \multicolumn{1}{c}{\begin{tabular}[c]{@{}c@{}}\modelName\\ :Dot-product\end{tabular}} \\ \midrule
$\mathbf{e}_{ij}^{(l)\prime} + \mathbf{e}_{ij, \text{CA}}^{(l)}$ &
  0.756${}_{\pm0.001}$ &
  0.757${}_{\pm0.001}$ \\
$\textsc{Concat} (\mathbf{e}_{ij}^{(0)}, \mathbf{e}_{ij, \text{CA}}^{(l)})$ &
  0.752${}_{\pm0.001}$ &
  0.752${}_{\pm0.001}$ \\ \bottomrule
\end{tabular}
\end{table}

In Sec.~\ref{sec:method} we introduce two ways to augment the computed content-attention (CA) edge embeddings $\mathbf{e}_{ij, \text{CA}}^{(l)}$ to edge embeddings $\mathbf{e}_{ij}^{(l)}$ between nodes $v_i$ and $v_j$ at $l^{th}$ layer.
The first method is to first update the edge embeddings without the content attention ($\mathbf{e}_{ij}^{(l)\prime}$) and use the element-wise addition.
The second method is to concatenate with the input edge feature $\mathbf{e}_{ij}^{(0)}$.
We compare the predictive performances of these methods for the Goodreads and Eedi datasets.
In Tables~\ref{tab:appendix_combination_goodreads} and \ref{tab:appendix_combination_eedi}, element-wise addition yields better predictive performance than concatenation for on both datasets with varying methods of attention computation: concat and dot-product.

\begin{table*}[t!]
\centering
\caption{Predictive performance of \modelName for bidirectional setting when concatenation is used for attention computation.}
\begin{tabular}{@{}lccrr@{}}
\toprule
\multicolumn{1}{c}{\textbf{Concat}} & Goodreads                         & \multicolumn{3}{c}{EEDI}                                                                 \\ \midrule
                                    & RMSE ($\downarrow$)                              & Accuracy ($\uparrow$)                          & \multicolumn{1}{c}{AUROC ($\uparrow$)} & \multicolumn{1}{c}{AUPR ($\uparrow$)} \\
User-only                           & \multicolumn{1}{r}{0.879 $\pm$ 0.000} & \multicolumn{1}{r}{0.756 $\pm$ 0.002} & 0.715 $\pm$ 0.002             & 0.874 $\pm$ 0.002            \\
\textbf{Bidirectional}              & \multicolumn{1}{r}{\textbf{0.873}$\pm$0.001} & \multicolumn{1}{r}{\textbf{0.761}$\pm$0.001} & \textbf{0.720}$\pm$0.002             & \textbf{0.891}$\pm$0.001            \\ \bottomrule
\end{tabular}
\label{tbl:bidirection1}
\end{table*}

\begin{table*}[t!]
\centering
\caption{Predictive performance of \modelName for bidirectional setting when dot-product is used for attention computation.}
\begin{tabular}{@{}lccrr@{}}
\toprule
\multicolumn{1}{c}{\textbf{Dot-product}} & Goodreads                         & \multicolumn{3}{c}{EEDI}                                                                 \\ \midrule
                                    & RMSE ($\downarrow$)                              & Accuracy ($\uparrow$)                          & \multicolumn{1}{c}{AUROC ($\uparrow$)} & \multicolumn{1}{c}{AUPR ($\uparrow$)} \\
User-only                           & \multicolumn{1}{r}{0.879 $\pm$ 0.001} & \multicolumn{1}{r}{0.757 $\pm$ 0.001} & 0.717 $\pm$ 0.001             & 0.874 $\pm$ 0.001            \\
\textbf{Bidirectional}              & \multicolumn{1}{r}{\textbf{0.872}$\pm$0.001} & \multicolumn{1}{r}{\textbf{0.760}$\pm$0.002} & \textbf{0.721}$\pm$0.002             & \textbf{0.888}$\pm$0.001            \\ \bottomrule
\end{tabular}
\label{tbl:bidirection2}
\end{table*}
\subsection{Bi-directional setting for \modelName}
In the formulation introduced in Eq.~\ref{eqn:edge_emb}, the computation of the content attention vector is skipped when the target node is an item and the source node is a user. In fact, \modelName works in more general settings than the recommendation system, where every node can potentially be associated with contents. However, in a recommendation system with bipartite graphs, it is common that only item nodes are associated with such content information. We have made a change in this formulation so that the update of the content attention (CA) vector is now bi-directional, e.g., whenever gets updated, update in the same fashion instead of skipping. In Tables~\ref{tbl:bidirection1} and~\ref{tbl:bidirection2}, we report results when the update of CA edge representation is bi-directional by simultaneously updating both from $i$ to $j$ and $j$ to $i$ edges. The results show the significant improvement in the predictive performance.
%

%

\section{Additional details on model training}
\subsection{Datasets}
\label{appendix:datasets}
Here, we provide additional information about the two real-world datasets used in our experiment.
We chose the Goodreads and Eedi datasets because they contains text information in sentences associated with each item. 

We use Goodreads dataset from~\citet{goodreads2020}.
We filtered out books whose descriptions are written in non-English languages,
and removed duplicate books based on their titles.
Originally, the ratings were text-based.
We converted the ratings as follows:
"Did not like it" to rating 1,
"It was okay" to rating 2,
"Liked it" to rating 3,
"Really liked it" to rating 4,
and "It was amazing" to rating 5.
We used Eedi dataset from~\citet{wang2020diagnostic}.
The content of the text information is extracted using optical character recognition (OCR) from the raw question images, as no question text is available.

In order to train \modelName and comparison models on a single GPU within our computation infrastructure, 
we took a subset of the Eedi dataset,
taking student responses from between March $4^{th}$ and March $27^{th}$.
\subsection{Configurations for the baseline methods}
\label{appendix:additional_config}
We detail the configurations used specific to each baseline for recording the test performance.
For GC-MC, we assign the separate message passing channels and their corresponding parameters for modeling different discrete edge labels.
The number of layer is set to 1, and We do not use the weight sharing method.
For the accumulation method, we use concatenation.
For GraphSAGE, we use the neighbor sampling size of 32 throughout all message passing layers.
For GRAPE, we do not use the one-hot node initializations for both Eedi and Goodreads, because of the large number of item nodes leading to GPU memory errors.
Instead, we use random initialization just like all GNN model configurations in our experiment.
For GAT, we use a single self-attention head. Alternatively, we also tested using multi-head attention with 4 heads with smaller $C^{(l,h)} = 16$, but the predictive performance did not increase. We could not test multi-head attention with $C^{(l,h)} = 64$ due to the GPU memory limits.
For GIN, we make the epsilon parameter trainable (Noted as GIN-$\epsilon$ in the original GIN paper). For JK, we choose LSTM during aggregation (Noted as JK-LSTM in the original JK paper). Both settings have shown to perform best amongst all configurations in the respective papers. 

\subsection{Computing infrastructure}
\label{appendix:computation}
Each experiment was run on a single GPU, which was either an NVIDIA Tesla K80 or an NVIDIA Tesla V100.
All experiments were scheduled and performed in Azure Machine Learning.